\documentclass[11pt,a4paper]{article}
\usepackage[hyperref]{emnlp2018}
\usepackage{times}
\usepackage{latexsym}
\usepackage{url}

\usepackage{amsmath}
\usepackage{arydshln}

\DeclareMathOperator{\Relu}{ReLU}

\usepackage{tikz}
\usetikzlibrary{shapes, shapes.geometric, arrows, cd, positioning, automata}
\tikzstyle{boxneuron} = [rectangle, minimum width=1.5cm, minimum height=1cm, text centered, draw=black]
\tikzstyle{arrow} = [thick,->,>=stealth]

\aclfinalcopy 

\usepackage{color}

\newcommand{\highlight}[1]{{\color{magenta} #1}}

\title{Context-Free Transductions with Neural Stacks}

\author{Yiding Hao,\thanks{\; Equal contribution.}\: William Merrill,\footnotemark[1]\: Dana Angluin, Robert Frank,\\ \textbf{Noah Amsel, Andrew Benz, \and Simon Mendelsohn} \\ 
	Department of Linguistics, Yale University \\
	Department of Computer Science, Yale University\\
	{\tt firstname.lastname@yale.edu}\\}

\date{}

\begin{document}
	
	\maketitle
	
	\begin{abstract}
		This paper analyzes the behavior of stack-augmented recurrent neural network (RNN) models. Due to the architectural similarity between stack RNNs and pushdown transducers, we train stack RNN models on a number of tasks, including string reversal, context-free language modelling, and cumulative XOR evaluation. Examining the behavior of our networks, we show that stack-augmented RNNs can discover intuitive stack-based strategies for solving our tasks. However, stack RNNs are more difficult to train than classical architectures such as LSTMs. Rather than employ stack-based strategies, more complex networks often find approximate solutions by using the stack as unstructured memory. 
	\end{abstract}
	
	\section{Introduction}
	
	Recent work on recurrent neural network (RNN) architectures has introduced a number of models that enhance traditional networks with differentiable implementations of common data structures. Appealing to their Turing-completeness \citep{siegelmannComputationalPowerNeural1995}, \citet{gravesNeuralTuringMachines2014} view RNNs as computational devices that learn transduction algorithms, and develop a trainable model of random-access memory that can simulate Turing machine computations. In the domain of natural language processing, the prevalence of context-free models of natural language syntax has motivated stack-based architectures such as those of \citet{grefenstetteLearningTransduceUnbounded2015} and \citet{joulinInferringAlgorithmicPatterns2015}. By analogy to \citeauthor{gravesNeuralTuringMachines2014}'s Neural Turing Machines, these stack-based models are designed to simulate pushdown transducer computations.
	
	From a practical standpoint, stack-based models may be seen as a way to optimize networks for discovering dependencies of a hierarchical nature. Additionally, stack-based models could potentially facilitate interpretability by imposing structure upon the recurrent state of an RNN. Classical architectures such as Simple RNNs \citep{elmanFindingStructureTime1990}, Long Short-Term Memory networks (LSTM, \citealp{hochreiterLongShortTermMemory1997}), and Gated Recurrent Unit networks (GRU, \citealp{choLearningPhraseRepresentations2014}) represent state as black-box vectors. In certain cases, these models can learn to implement classical data structures using state vectors \citep{kirov2011}. However, because state vectors are fixed in size, the inferred data structures must be represented in a fractal encoding requiring arbitrary position. On the other hand, differentiable stacks typically increase in size throughout the course of the computation, so their performance may better scale to larger inputs. Since the ability of a differentiable stack to function correctly intrinsically requires that the information it contains be represented in the proper format, examining the contents of a network's stack throughout the course of its computation could reveal hierarchical patterns that the network has discovered in its training data. 
	
	This paper systematically explores the behavior of stack-augmented RNNs on simple computational tasks. While \citet{yogatamaMemoryArchitecturesRecurrent2018} provide an analysis of stack RNNs based on their Multipop Adaptive Computation Stack model, our analysis is based on the existing Neural Stack model of \citet{grefenstetteLearningTransduceUnbounded2015}, as well as a novel enhancement thereof. We consider tasks with optimal strategies requiring either finite-state memory or a stack, or possibly a combination of the two. We show that Neural Stack networks have the ability to learn to use the stack in an intuitive manner. However, we find that Neural Stacks are more difficult to train than classical architectures. In particular, our models prefer not to employ stack-based strategies when other forms of memory are available, such as in networks with both LSTM memory and a stack. 
	
	A description of our models, including a review of \citeauthor{grefenstetteLearningTransduceUnbounded2015}'s Neural Stacks, appears in Section 2. Section 3 discusses the relationship between stack-augmented RNN models and pushdown transducers, motivating our intuition that Neural Stacks are a suitable architecture for learning context-free structure. The tasks we consider are defined in Section 4, and our experimental paradigm is described in Section 5. Section 6 presents quantitative evaluation of our models' performance as well as qualitative description of their behavior. Section 7 concludes.
	
	\section{Models}
	\begin{figure}
		\begin{center}
			\begin{tikzpicture}[scale=1.0, every node/.style={scale=.9}]
			
			\node (controller) [boxneuron] {Controller};
			\node (stack) [boxneuron, right of=controller, xshift=3cm] {Stack};
			\node (prevstack) [below of=stack, yshift=-.5cm] {$\langle \mathbf{V}_{t - 1}, \mathbf{s}_{t - 1} \rangle$};
			
			\node (input) [below of=controller, yshift=-.5cm] {$\langle \mathbf{x}_t, \mathbf{h}_{t - 1}, \mathbf{r}_{t - 1} \rangle$};
			\node (output) [above of=controller, yshift=.5cm] {$\langle \mathbf{y}_t, \mathbf{h}_t \rangle$};
			
			\node (read) [above of=stack, yshift=.5cm] {$\langle \mathbf{r}_t, \mathbf{V}_t, \mathbf{s}_t \rangle$};
			
			\coordinate [right of=controller, xshift=1.75cm] (n1);

			\draw [arrow] (controller) -- node[below] {$\langle \mathbf{v}_t, u_t, d_t \rangle$} (n1) -- (stack);
			\draw [arrow] (input) -- (controller);
			\draw [arrow] (controller) -- (output);
			\draw [arrow] (stack) -- (read);
			\draw [arrow] (prevstack) -- (stack);
			
			\end{tikzpicture}
			
			\caption{The Neural Stack architecture.}
		\end{center}
	\end{figure}
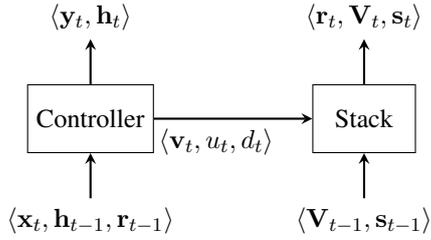
	
	The neural network models considered in this paper are based on the Neural Stacks of \citet{grefenstetteLearningTransduceUnbounded2015}, a family of stack-augmented RNN architectures.\footnote{Code for our PyTorch \cite{paszke2017automatic} implementation is available at \url{https://github.com/viking-sudo-rm/StackNN}.} A Neural Stack model consists of two modular components: a \textit{controller} executing the computation of the network and a \textit{stack} implementing the data model of the network. At each time step $t$, the controller receives an input vector $\mathbf{x}_t$ and a \textit{read vector} $\mathbf{r}_{t - 1}$ representing the material at the top of the stack at the end of the previous time step. We assume that the controller may adhere to any feedforward or recurrent structure; if the controller is recurrent, then it may also receive a recurrent state vector $\mathbf{h}_{t - 1}$. Based on $\mathbf{x}_t$, $\mathbf{r}_{t - 1}$, and possibly $\mathbf{h}_{t - 1}$, the controller computes an output $\mathbf{y}_t$, a new recurrent state vector $\mathbf{h}_t$ if applicable, and a tuple $\langle \mathbf{v}_t, u_t, d_t \rangle$ containing instructions for manipulating the stack. The stack takes these instructions and produces $\mathbf{r}_t$, the vector corresponding to the material at the top of the stack after popping and pushing operations have been performed on the basis of $\langle \mathbf{v}_t, u_t, d_t \rangle$. The contents of the stack are represented by a recurrent state matrix $\mathbf{V}_t$ and a strength vector $\mathbf{s}_t$. This schema is shown in Figure 1.
	
	Having established the basic architecture, the remainder of this section introduces our models in full detail. Subsection 2.1 describes how the stack computes $\mathbf{r}_t$ and updates $\mathbf{V}_t$ and $\mathbf{s}_t$ based on $\langle \mathbf{v}_t, u_t, d_t \rangle$. Subsection 2.2 presents the various kinds of controllers we consider in this paper. Subsection 2.3 presents an enhancement of \citeauthor{grefenstetteLearningTransduceUnbounded2015}'s schema that allows the network to perform computations of varying duration.
	
	\subsection{Differentiable Stacks}
	
	A stack at time $t$ consists of sequence of vectors $\langle \mathbf{V}_t[1], \mathbf{V}_t[2], \dots, \mathbf{V}_t[t] \rangle$, organized into a matrix $\mathbf{V}_t$ whose $i$th row is $\mathbf{V}_t[i]$. By convention, $\mathbf{V}_t[t]$ is the ``top'' element of the stack, while $\mathbf{V}_t[1]$ is the ``bottom'' element. Each element $\mathbf{V}_t[i]$ of the stack is associated with a \textit{strength} $\mathbf{s}_t[i] \in [0, 1]$. The strength of a vector $\mathbf{V}_t[i]$ represents the ``degree'' to which the vector is on the stack: a strength of $1$ means that the vector is ``fully'' on the stack, while a strength of $0$ means that the vector has been popped from the stack. The strengths are organized into a vector $\mathbf{s}_t = \langle \mathbf{s}_t[1], \mathbf{s}_t[2], \dots, \mathbf{s}_t[t] \rangle$.
	
	At each time step, the stack \textit{pops} a number of items from the top, \textit{pushes} a new item to the top, and \textit{reads} a number of items from the top, in that order. The behavior of the popping and pushing operations is determined by the instructions $\langle \mathbf{v}_t, u_t, d_t \rangle$. The value obtained from the reading operation is passed back to the controller as the recurrent vector $\mathbf{r}_t$. Let us now describe each of the three operations.
	
	Popping reduces the strength $\mathbf{s}_{t - 1}[t - 1]$ of the top element from the previous time step by $u_t$. If $\mathbf{s}_{t - 1}[t - 1] \geq u_t$, then the strength of the $(t - 1)$st element after popping is simply $\mathbf{s}_t[t - 1] = \mathbf{s}_{t - 1}[t - 1] - u_t$. If $\mathbf{s}_{t - 1}[t - 1] \leq u_t$, then we consider the popping operation to have ``consumed'' $\mathbf{s}_{t - 1}[t - 1]$, and the strength $\mathbf{s}_{t - 1}[t - 2]$ of the next element is reduced by the ``left-over'' strength $u_t - \mathbf{s}_{t - 1}[t - 1]$. This process is repeated until all strengths in $\mathbf{s}_{t - 1}$ have been reduced. For each $i < t$, we compute the left-over popping strength $\mathbf{u}_t[i]$ for the $i$th item as follows.
	\begin{align*}
	&\mathbf{u}_t[i] = \\
	&\begin{cases}
	u_t, & i = t - 1 \\
	\Relu(\mathbf{u}_t[i + 1] - \mathbf{s}_{t - 1}[i + 1]), & i < t - 1
	\end{cases}
	\end{align*}
	The strengths are then updated accordingly.
	\[
	\mathbf{s}_t[i] = \Relu\left(\mathbf{s}_{t - 1}[i] -  \mathbf{u}_t[i] \right)
	\]
	
	The pushing operation simply places the vector $\mathbf{v}_t$ at the top of the stack with strength $d_t$. Thus, $\mathbf{V}_t$ and $\mathbf{s}_t[t]$ are updated as follows.
	\[
    \mathbf{s}_t[t] = d_t
    \qquad
    \mathbf{V}_t[i] = \begin{cases}
	\mathbf{v}_t, & i = t \\
	\mathbf{V}_{t - 1}[i], & i < t
	\end{cases} 
	\]
	Note that $\mathbf{s}_t[1]$, $\mathbf{s}_t[2]$, \dots, $\mathbf{s}_t[t - 1]$ have already been updated during the popping step.
	
	The reading operation ``reads'' the elements on the top of the stack whose total strength is $1$. If $\mathbf{s}_t[t] = 1$, then only the top element is read. Otherwise, the next element is read using the ``left-over'' strength $1 - \mathbf{s}_t[t]$. As in the case of popping, we may define a series of left-over strengths $\rho_t[1]$, $\rho_t[2]$, \dots, $\rho_t[t]$ corresponding to each item in the stack.
	\[
	\rho_t[i] = \begin{cases}
	1, & i = t \\
	\Relu\left( \rho_t[i + 1] - \mathbf{s}_t[i + 1] \right), & i < t
	\end{cases}
	\]
	The result $\mathbf{r}_t$ of the reading operation is obtained by computing a sum of the items in the stack weighted by their strengths, including only items with sufficient left-over strength.
	\[
	\mathbf{r}_t = \sum_{i = 1}^t \min\left(\mathbf{s}_t[i], \rho_t[i] \right) \cdot \mathbf{V}_t[i]
	\]
	
	\subsection{Controllers}
	\newcommand{\XR}{\left[ \begin{array}{c;{2pt/2pt}c}
			\mathbf{x}_t & \mathbf{r}_{t - 1}
		\end{array} \right]^\top}
	\newcommand{\XRH}{\left[ \begin{array}{c;{2pt/2pt}c;{2pt/2pt}c}
			\mathbf{x}_t & \mathbf{r}_{t - 1} & \mathbf{h}_{t - 1}
		\end{array} \right]^\top}
	
	We consider two types of controllers: \textit{linear} and \textit{LSTM}. The linear controller is a feedforward network consisting of a single linear layer. The network output is directly extracted from the linear layer, while the stack instructions are passed through the sigmoid function, denoted $\sigma$.
	\begin{align*}
	u_t &= \sigma\left( \mathbf{W}_u \cdot \XR + b_u \right)\\
	d_t &= \sigma\left( \mathbf{W}_d \cdot \XR + b_d \right)\\
	\mathbf{v}_t &= \sigma\left( \mathbf{W}_{\mathbf{v}} \cdot \XR + \mathbf{b}_{\mathbf{v}} \right)\\ 
	\mathbf{y}_t &=  \mathbf{W}_{\mathbf{y}} \cdot \XR + \mathbf{b}_{\mathbf{y}}
	\end{align*}
	The LSTM controller maintains two state vectors: the \textit{hidden state} $\mathbf{h}_t$ and the \textit{cell state} $\mathbf{c}_t$. The output and stack instructions are produced by passing $\mathbf{h}_t$ through a linear layer. As in the linear controller, the stack instructions are additionally passed through the sigmoid function.
	
	\subsection{Buffered Networks}
	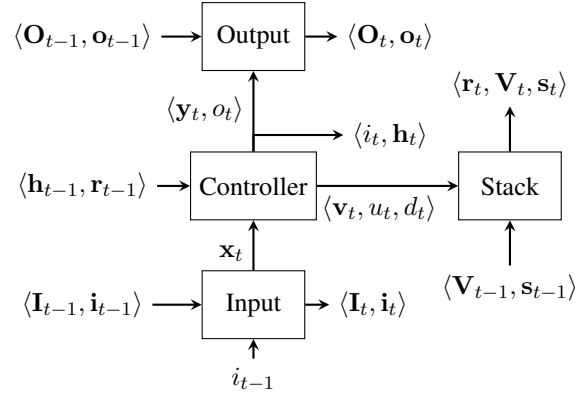
\begin{figure}
		\begin{center}
			\begin{tikzpicture}[scale=.9, every node/.style={scale=.9}]
			
			\node (controller) [boxneuron] {Controller};
			\node (stack) [boxneuron, right of=controller, xshift=2.75cm] {Stack};
			\node (prevstack) [below of=stack, yshift=-.5cm] {$\langle \mathbf{V}_{t - 1}, \mathbf{s}_{t - 1} \rangle$};
			
			\coordinate [above of=controller, yshift=-.25cm] (n3);
			\coordinate [above of=n3, yshift=-.25cm] (n4);
			\coordinate [below of=controller, yshift=.25cm] (n2);
			
			\node (bufferin) [boxneuron, below of=n2] {Input};
			\node (bufferout) [boxneuron, above of=n4, yshift=-.3cm] {Output};
			\node (prevbufferout) [left of=bufferout, xshift=-1.5cm] {$\langle \mathbf{O}_{t - 1}, \mathbf{o}_{t - 1} \rangle$};
			\node (nextbufferin) [right of=bufferin, xshift=.75cm] {$\langle \mathbf{I}_t, \mathbf{i}_t \rangle$};
			
			\node (input) [left of=bufferin, xshift=-1.5cm] {$\langle \mathbf{I}_{t - 1}, \mathbf{i}_{t - 1} \rangle$};
			\node (output) [right of=bufferout, xshift=1cm] {$\langle  \mathbf{O}_t, \mathbf{o}_t \rangle$};
			
			\node (read) [above of=stack, yshift=.5cm] {$\langle \mathbf{r}_t, \mathbf{V}_t, \mathbf{s}_t \rangle$};
			\node (statein) [left of=controller, xshift=-1.5cm] {$\langle \mathbf{h}_{t - 1}, \mathbf{r}_{t - 1} \rangle$};
			\node (stateout) [right of=n3, xshift=1cm] {$\langle i_t, \mathbf{h}_t \rangle$};
			
			\node (ein) [below of=bufferin, yshift=-.1cm] {$i_{t - 1}$};
			
			\coordinate [right of=controller, xshift=1.75cm] (n1);

			\draw [arrow] (controller) -- node [below] {$\langle \mathbf{v}_t, u_t, d_t \rangle$} (n1) -- (stack);
			
			\draw [arrow] (bufferin) -- node [left] {$\mathbf{x}_t$} (n2) -- (controller);
			
			\draw [arrow] (controller) -- (n3) -- node [left] {$\langle \mathbf{y}_t, o_t \rangle$} (n4) -- (bufferout);
			
			\draw [arrow] (bufferout) -- (output);
			
			\draw [arrow] (input) -- (bufferin);
			\draw [arrow] (statein) -- (controller);
			\draw [arrow] (controller) -- (n3) -- (stateout);
			\draw [arrow] (stack) -- (read);
			\draw [arrow] (prevbufferout) -- (bufferout);
			\draw [arrow] (bufferin) -- (nextbufferin);
			\draw [arrow] (prevstack) -- (stack);
			\draw [arrow] (ein) -- (bufferin);
			\end{tikzpicture}
			
			\caption{Our enhanced architecture with buffers.}
		\end{center}
	\end{figure}
	
	One limitation of many RNN architectures, including Neural Stacks, is that they can only compute \textit{same-length} transductions: at each time step, the network must accept exactly one input vector and produce exactly one output vector. This limitation prevents Neural Stacks from producing output sequences that may be longer or shorter than the input sequence. It also prohibits Neural Stack networks from performing computation steps without reading an input or producing an output (i.e., $\varepsilon$-transitions on input or output), even though such computation steps are a common feature of stack transduction algorithms.
	
	A well-known approach to overcoming this limitation appears in Sequence-to-Sequence models such as \citet{sutskeverSequenceSequenceLearning2014} and \citet{choLearningPhraseRepresentations2014}. There, the production of the output sequence is delayed until the input sequence has been fully read by the network. Output vectors produced while reading the input are discarded, and the input sequence is padded with blank symbols to indicate that the network should be producing an output. 
	
	The delayed output approach solves the problem of fixed-length outputs, and we adopt it for the String Reversal task described in Section 4. However, delaying the output does not allow our networks to perform streaming computations that may interrupt the process of reading inputs or emitting outputs. An alternative approach is to allow our networks to perform $\varepsilon$-transitions. While \citet{gravesAdaptiveComputationTime2016} achieves this by dynamically repeating inputs and marking them with flags, we augment the Neural Stack architecture with two differentiable buffers: a read-only input buffer and a write-only output buffer. At each time step $t$, the input vector $\mathbf{x}_t$ is obtained by popping from the input buffer with strength $i_{t - 1}$. In addition to the output vector and stack instructions, the controller must produce an input buffer pop strength $i_t$ and an output buffer push strength $o_t$. The output vector is then enqueued to the output buffer with strength $o_t$. This enhanced architecture is shown in Figure 2.
	
	The implementation of the input and output buffers is based on \citet{grefenstetteLearningTransduceUnbounded2015}'s Neural Queues, a first-in-first-out variant of the Neural Stack. Like the stack, the input buffer at time $t$ consists of a matrix of vectors $\mathbf{I}_t$ and a vector of strengths $\mathbf{i}_t$. Similarly, the output buffer consists of a matrix of vectors $\mathbf{O}_t$ and a vector of strengths $\mathbf{o}_t$. The input buffer is initialized so that $\mathbf{I}_0$ is a matrix representation of the full input sequence, with an initial strength of $1$ for each item. 
	
	At time $t$, items are dequeued from the ``front'' of the buffer with strength $i_{t - 1}$.
	\begin{align*}
	\iota_t[j] &= \begin{cases}
	i_{t - 1}, & j = 1 \\
	\Relu(\iota_t[j - 1] - \mathbf{i}_{t - 1}[j]), & j > 1
	\end{cases} \\
	\mathbf{i}_t[j] &= \Relu\left(\mathbf{i}_{t - 1}[j] - \iota_t[j]\right)
	\end{align*}
	Next, the input vector $\mathbf{x}_t$ is produced by reading from the front of the buffer with strength $1$.
	\begin{align*}
	\xi_t[j] &= \begin{cases}
	1, & j = 1 \\
	\Relu\left( \xi_t[j - 1] - \mathbf{i}_t[j] \right), & j > 1
	\end{cases} \\
	\mathbf{x}_t &= \sum_{j = 1}^n \min\left(\mathbf{i}_t[j], \xi_t[j] \right) \cdot \mathbf{I}_t[j]
	\end{align*}
	Since the input buffer is read-only, there is no push operation. This means that unlike $\mathbf{V}_t$ and $\mathbf{O}_t$, the number of rows of $\mathbf{I}_t$ is fixed to a constant $n$. When the controller's computation is complete, the output vector $\mathbf{y}_t$ is enqueued to the ``back'' of the output buffer with strength $o_t$.
	\begin{align*}
	\mathbf{O}_t[j] &= \begin{cases}
	\mathbf{y}_t, & j = t \\
	\mathbf{O}_{t - 1}[j], & j < t
	\end{cases} \\
	\mathbf{o}_t[j] &= \begin{cases}
	o_t, & j = t \\
	\mathbf{o}_{t - 1}[j], & j < t
	\end{cases}
	\end{align*}
	After the last time step, the final output sequence is obtained by repeatedly dequeuing the front of the output buffer with strength $1$ and reading the front of the output with strength $1$. These dequeuing and reading operations are identical to those defined for the input buffer.
	
	\section{Pushdown Transducers}
	
	Our decision to use a stack for NLP tasks rather than some other differentiable data structure is motivated by the success of context-free grammars (CFGs) in describing the hierarchical phrase structure of natural language syntax. A classic theoretical result due to \citet{chomskyContextfreeGrammarsPushdown1962} shows that CFGs generate exactly those sets of strings that are accepted by nondetermininstic pushdown automata (PDAs), a model of computation that augments a finite-state machine with a stack. When enhanced with input and output buffers, we consider Neural Stacks to be an implementation of \textit{deterministic pushdown transducers} (PDTs), a variant of PDAs that includes an output tape.
	
	Formally, a PDT is described by a \textit{transition function} of the form $\delta(q, x, s) = \left\langle q^\prime, y, s^\prime \right\rangle$, interpreted as follows: if the machine receives an $x$ from the input buffer and pops an $s$ from the top of the stack while in state $q$, then it sends a $y$ to the output buffer, pushes an $s^\prime$ to the stack, and transitions to state $q^\prime$. We assume that $\delta$ is only defined for finitely many configurations $\langle q, x, s \rangle$. These configurations, combined with their corresponding values of $\delta$, represent all the possible actions of a pushdown transducer.
	
	To illustrate, let us construct a PDT that computes the function $f\left(w\#^{|w|}\right) = \#^{|w|}w^R,$ where $w^R$ is the reverse of $w$ and $\#^{|w|}$ is a sequence of $\#$s of the same length as $w$. We can begin to compute $f$ using a single state $q_0$ by pushing each symbol of $w$ onto the stack while emitting $\#$s as output. When the machine has finished reading $w$, the stack contains the symbols of $w$ in reverse order. In the remainder of the computation, the machine pops symbols from the stack one at a time and sends them to the output buffer. A pictoral representation of this PDT is shown in Figure 3. Each circle represents a state of the PDT, and each action $\delta(q, x, s) = \langle q^\prime, y, s^\prime \rangle$ is represented by an arrow from $q$ to $q^\prime$ with the label ``$x:y, s \to s^\prime$.'' Observe that the two labels of the arrow from $q_0$ to itself encode a transition function implementing the algorithm described above.
	\begin{figure}
		\begin{center}
			\begin{tikzpicture}[>=stealth',shorten >=1pt,auto,node distance=2cm]
			
			\node[initial,state,accepting]  (q1) {$q_0$};
			
			\path[->] (q1) edge [loop right] node 
			{\parbox{3cm}{$x:\#, \varepsilon \to x$\\$\#:y, y \to \varepsilon$}} (q1);
			\end{tikzpicture}
			\caption{A PDT for the String Reversal task.}
		\end{center}
	\end{figure}
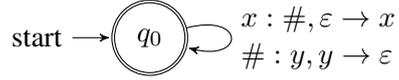
	
	Given a finite state transition function, there exists an LSTM that implements it. In fact, \citet{weissPracticalComputationalPower2018} show that a deterministic $k$-counter automaton can be simulated by an LSTM. Thus, any deterministic PDT can be simulated by the buffered stack architecture with an LSTM controller.
	
	\section{Tasks}
	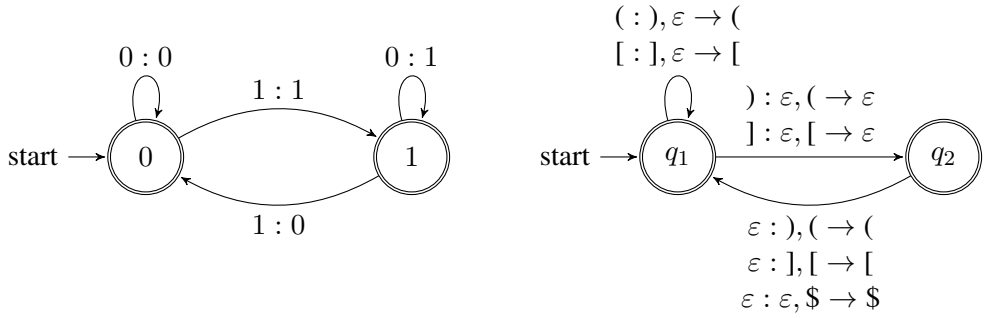
\begin{figure*}
		\begin{center}
			\begin{tikzpicture}[>=stealth',shorten >=1pt,auto,node distance=2cm]
			
			\node[initial,state,accepting]  (q1) {$q_1$};
			\node[state, accepting]         (q2) [right of=q1, xshift=1.5cm] {$q_2$};
			
			\path[->] (q1) edge [loop above] node 
			{\parbox{2cm}{\centering $\text{(}:\text{)}, \varepsilon \to \text{(}$\\$\text{[}:\text{]},
					\varepsilon \to \text{[}$}} (q1)

			(q1) edge node 
			{\parbox{2cm}{\centering $\text{)}:\varepsilon, \text{(} \to \varepsilon$\\$\text{]}:\varepsilon,  \text{[} \to \varepsilon$}} (q2)

			(q2) edge [bend left] node 
			{\parbox{2cm}{\centering $\varepsilon:\text{)}, \text{(} \to \text{(}$\\$\varepsilon:\text{]}, \text{[} \to \text{[}$\\$\varepsilon:\varepsilon,\text{\$}\to \text{\$}$}} (q1);

			\node[state, accepting]         (q4) [left of=q1, xshift=-1.5cm] {$1$};
			\node[initial,state,accepting]  (q3) [left of=q4, xshift=-1.5cm] {$0$};
			
			\path[->] (q3) edge [loop above] node 
			{$0:0$} (q3)
			
			(q3) edge [bend left] node 
			{$1:1$} (q4)
			
			(q4) edge [loop above] node 
			{$0:1$} (q4)
			
			(q4) edge [bend left] node 
			{$1:0$} (q3);
			\end{tikzpicture}

			\caption{PDTs for Cumulative XOR Evaluation (left) and Parenthesis Prediction (right) tasks. The symbol \$\ represents the bottom of the stack.}
		\end{center}
	\end{figure*}
	
	The goal of this paper is to ascertain whether or not stack-augmented RNN architectures can learn to perform PDT computations. To that end, we consider six tasks designed to highlight various features of PDT algorithms. Four of these tasks---String Reversal, Parenthesis Prediction, and the two XOR Evaluation tasks---have simple PDT implementations. The PDTs for each of these tasks differ in their memory requirements: they require either finite-state memory or stack-structured memory, or a combination of the two. The remaining two tasks---Boolean Formula Evaluation and Subject--Auxiliary Agreement---are designed to determine whether or not Neural Stacks can be applied to complex use cases that are thought to be compatible with stack-based techniques.
	
	\subsection{String Reversal}
	
	In the String Reversal task, the network must compute the function $f$ from the previous section. As discussed there, the String Reversal task can be performed straightforwardly by pushing all input symbols to the stack and then popping all symbols from the stack. The purpose of this task is to serve as a baseline test for whether or not a controller can learn to use a stack in principle. Since in the general case, correctly producing $w^R$ requires recording $w$ in the stack, we evaluate the network solely based on the portion of its output where $w^R$ should appear, immediately after reading the last symbol of $w$.
	
	\subsection{XOR Evaluation}
	
	We consider two tasks that require the network to implement the XOR function. In the \textit{Cumulative XOR Evaluation} task, the network reads an input string of $1$s and $0$s. At each time step, the network must output the XOR of all the input symbols it has seen so far. The \textit{Delayed XOR Evaluation} task is similar, except that the most recent input symbol is excluded from the XOR computation.
	
	As shown in the left of Figure 4, the XOR Evaluation tasks can be computed by a PDT without using the stack. Thus, we use XOR Evaluation to test the versatility of the stack by assessing whether a feedforward controller can learn to use it as unstructured memory.
	
	The Cumulative XOR Evaluation task presents the linear controller with a theoretical challenge because single-layer linear networks cannot compute the XOR function \citep{minskyPerceptronsIntroductionComputational1969}. However, in the Delayed XOR Evaluation task, the delay between reading an input symbol and incorporating it into the XOR gives the network two linear layers to compute XOR when unravelled through time. Therefore, we expect that the linear model should be able to perform the Delayed XOR Evaluation task, but not the Cumulative XOR Evaluation task.
	
	The discrepancy between the Cumulative and the Delayed XOR Evaluation tasks for the linear controller highlights the importance of timing in stack algorithms. Since the our enhanced architecture from Subsection 2.3 can perform $\varepsilon$-transitions, we expect it to perform the Cumulative XOR Evaluation task with a linear controller by learning to introduce the necessary delay. Thus, the XOR tasks allow us to test whether our buffered model can learn to optimize the timing of its computation.

	\subsection{Parenthesis Prediction}
	
	The Parenthesis Prediction task is a simplified language modelling task. At each time step $t$, the network reads the $t$th symbol of some string and must attempt to output the $(t + 1)$st symbol. The strings are sequences of well-nested parentheses generated by the following CFG.
	 \begin{align*}
       \text{S} &\to \text{S T} \mathrel{|} \text{T S} \mathrel{|} \text{T}\\
       \text{T} &\to \text{( T )} \mathrel{|} \text{( )} \\
       \text{T} &\to \text{[ T ]} \mathrel{|} \text{[ ]} 
   \end{align*}
   We evaluate the network only when the correct prediction is ) or ]. This restriction allows for a deterministic PDT solution, shown in the right of Figure 4. 
	
	Unlike String Reversal and XOR Evaluation, the Parenthesis Prediction task relies on both the stack and the finite-state control. Thus, the Parenthesis Prediction task tests whether or not Neural Stack models can learn to combine different types of memory. Furthermore, since context-free languages can be canonically represented as homomorphic images of well-nested parentheses \citep{chomskyAlgebraicTheoryContextFree1959}, the Parenthesis Prediction task may be used to gauge the suitability of Neural Stacks for context-free language modelling.
	
	\subsection{Boolean Formula Evaluation}
	
	In the Boolean Formula Evaluation task, the network reads a boolean formula in reverse Polish notation generated by the following CFG.
	   \begin{align*}
       \text{S} &\to \text{S S } \vee \mathrel{|} \text{S S }  \wedge\\
       \text{S} &\to \text{T} \mathrel{|} \text{F} 
   \end{align*} 
   At each time step, the network must output the truth value of the longest sub-formula ending at the input symbol.
	
	The Boolean Formula Evaluation task tests the ability of Neural Stacks to infer complex computations over the stack. In this case, the network must store previously computed values on the stack and evaluate boolean operations over these stored values. This technique is reminiscent of shift-reduce parsing, making the Boolean Formula Evaluation task a testing ground for the possibility of applying Neural Stacks to natural language parsing.
	
	\subsection{Subject--Auxiliary Agreement}
	
	The Subject--Auxiliary Agreement task is inspired by \citet{linzenAssessingAbilityLSTMs2016}, who investigate whether or not LSTMs can learn structure-sensitive long-distance dependencies in natural language syntax. There, the authors train LSTM models that perform language modelling on prefixes of sentences drawn from corpora. The last word of each prefix is a verb, and the models are evaluated solely on whether or not they prefer the correct form of the verb over the incorrect ones. In sentences with embedded clauses, the network must be able to identify the subject of the verb among several possible candidates in order to conjugate the verb.
	
	Here, we consider sentences generated by a small, unambiguous CFG that models a fragment of English. 
	  \begin{align*}
        \text{S} &\to \text{NPsing has} \mathrel{|} \text{NPplur have} \\
        \text{NP} &\to \text{NPsing} \mathrel{|} \text{NPplur} \\
        \text{NPsing} &\to \text{the lobster } (\text{PP}\mathrel{|}\text{Relsing}) \\
        \text{NPplur} &\to \text{the lobsters } (\text{PP}\mathrel{|}\text{Relplur}) \\
        \text{PP}  &\to \text{in NP} \\
        \text{Relsing}  &\to \text{that has VP} \mathrel{|} \text{Relobj} \\
        \text{Relplur}  &\to \text{that have VP} \mathrel{|} \text{Relobj}\\
        \text{Relobj}  &\to \text{that NPsing has devoured} \\
        \text{Relobj} &\to \text{that NPplur have devoured} \\
        \text{VP}  &\to \text{slept} \mathrel{|} \text{devoured NP} \\
    \end{align*}
    As in the Parenthesis Prediction task, the network performs language modelling, but is only evaluated when the correct prediction is an auxiliary verb (i.e., {\em has} or {\em have}).
	
	\section{Experiments}
	
	We conducted four experiments designed to assess various aspects of the behavior of Neural Stacks. In each experiment, models are trained on a generated dataset consisting of 800 input--output string pairings encoded in one-hot representation. Training occurs in mini-batches containing 10 string pairings each. At the end of each epoch, the model is evaluated on a generated development set of 100 examples. Training terminates when five consecutive epochs fail to exceed the highest development accuracy attained. The sizes of the LSTM controllers' recurrent state vectors are fixed to 10, and, with the exception of Experiment 2 described below, the sizes of the vectors placed on the stack are fixed to 2. After training is complete, each trained model is evaluated on a testing set of 1000 generated strings, each of which is at least roughly twice as long as the strings used for training. 10 trials are performed for each set of experimental conditions.
	
	Experiment 1 tests the propensity of trained Neural Stack models to use the stack. We train both the standard Neural Stack model and our enhanced buffered model from Subsection 2.3 to perform the String Reversal task using the linear controller. To compare the stack with unstructured memory, we also train the standard Neural Stack model using the LSTM controller as well as an LSTM model without a stack. Training and development data are obtained from sequences of $0$s and $1$s randomly generated with an average length of 10. The testing data have an average length of 20.
	
	\begin{table*}[h]
		\centering
		\begin{tabular}{lccc|ccc|ccc} \hline
			\textbf{Task} & \textbf{Buffered} & \textbf{Controller} & \textbf{Stack} & \textbf{Min} & \textbf{Med} & \textbf{Max} & \textbf{Min} & \textbf{Med} & \textbf{Max}  \\\hline
			
			Reversal & No & Linear & Yes & 49.9 & \textbf{100.0} & \textbf{100.0} & 49.3 & \textbf{100.0} & \textbf{100.0} \\
			
			Reversal & Yes & Linear & Yes & 55.3 & 98.7 & 99.4 & 49.5 & 60.4 & 74.7 \\
			
			Reversal & No & LSTM & Yes & 81.2 & 89.3 & 94.4 & \textbf{67.2} & 71.0 & 73.7 \\
			
			Reversal & No & LSTM & No & \textbf{83.0} & 86.5 & 92.5 & 64.8 & 68.6 & 73.3 \\\hline
			
			XOR & No & Linear & Yes & 51.1 & 53.5 & 54.4 & 50.7 & 51.9 & 51.9 \\
			
			XOR & No & LSTM & Yes & \textbf{100.0} & \textbf{100.0} & \textbf{100.0} & \textbf{99.7} & \textbf{100.0} & \textbf{100.0} \\
			
			XOR & Yes & Linear & Yes & 51.0 & 99.8 & \textbf{100.0} & 50.4 & 96.0 & 99.1 \\
			
			Delayed XOR & No & Linear & Yes & \textbf{100.0} & \textbf{100.0} & \textbf{100.0} & \textbf{100.0} & \textbf{100.0} & \textbf{100.0} \\\hline
			
			Parenthesis & No & Linear & Yes & 72.8 & 97.0 & 99.3 & 59.9 & 80.3 & 83.2 \\
			
			Parenthesis & No & Linear & No & 70.0 & 71.8 & 73.3 & 59.9 & 60.5 & 60.7 \\
			
			Parenthesis & No & LSTM & Yes & \textbf{100.0} & \textbf{100.0} & \textbf{100.0} & \textbf{85.8} & \textbf{86.8} & \textbf{88.9} \\
			
			Parenthesis & No & LSTM & No & \textbf{100.0} & \textbf{100.0} & \textbf{100.0} & 83.5 & 85.8 & 88.0 \\\hline
			
			Formula & No & Linear & Yes & 87.4 & 92.0 & 97.3 & 87.8 & 91.2 & 96.2 \\
			
			Formula & No & LSTM & Yes & \textbf{98.0} & \textbf{98.7} & \textbf{99.4} & \textbf{96.8} & \textbf{97.7} & \textbf{98.4} \\
			
			Formula & No & LSTM & No & 95.4 & 98.5 & 99.3 & 95.3 & 97.6 & \textbf{98.4} \\
			
			Agreement & No & Linear & Yes & 53.3 & 73.5 & 93.9 & 51.8 & 68.8 & 85.8 \\
			
			Agreement & No & LSTM & Yes & 95.6 &  \textbf{98.5} & 99.7 & 82.4 & \textbf{88.8} & \textbf{91.2} \\
			
			Agreement & No & LSTM & No & \textbf{96.2} & 98.1 & \textbf{100.0} & \textbf{83.7} & 88.2 & 90.6 \\\hline
		\end{tabular}
		\caption{The minimum, median, and maximum accuracy (\%) attained by the 10 models for each experimental condition during the last epoch of the training phase (left) and the final testing phase (right).}
		\label{tab:my_label}
	\end{table*}
	
	Experiment 2 considers the XOR Evaluation tasks. We train standard models with a linear controller on the Delayed XOR task and an LSTM controller on the Cumulative XOR task to test the network's ability to use the stack as unstructured state. We also train both a standard and a buffered model on the Cumulative XOR Evaluation task using the linear controller to test the network's ability to use our buffering mechanism to infer optimal timing for computation steps. Training and development data are obtained from randomly generated sequences of $0$s and $1$s fixed to a length of 12. The testing data are fixed to a length of 24. The vectors placed on the stack are fixed to a size of 6.

	In Experiment 3, we attempt to perform the Parenthesis Prediction task using standard models with various types of memory: a linear controller with no stack, which has no memory; a linear controller with a stack, which has stack-structured memory; an LSTM controller with no stack, which has unstructured memory; and an LSTM controller with a stack, which has both stack-structured and unstructured memory.
	
	Sequences of well-nested parentheses are generated by the CFG from the previous section. The training and development data are obtained by randomly sampling from the set of strings of derivation depth at most 6, which contains strings of length up to 20. The testing data are of depth 12 and length up to 110.
	
	Experiment 4 compares the standard models with linear and LSTM controllers against a baseline consisting of an LSTM controller with no stack. Whereas Experiments 1--3 presented the network with tasks designed to showcase various features of the Neural Stack architecture, the goal of this experiment is to gauge the extent to which stack-structured memory may improve the network's performance on more sophisticated tasks. We train the three types of models on the Boolean Formula Evaluation task and the Subject--Auxiliary Agreement task. Data for both tasks are generated by the CFGs given in Section 4. The boolean formulae for training and development are randomly sampled from the set of strings of derivation depth at most 6, having a maximum length of 15, while the testing data are sampled from derivations of depth at most 7, with a maximum length of 31. The sentence prefixes are of depth 16 and maximum length 23 during the training phase, and depth 32 and maximum length 49 during the final evaluation round.

	\begin{figure*}[h]
		\begin{center}
			\begin{tabular}{ccc}
				\includegraphics[scale=0.3]{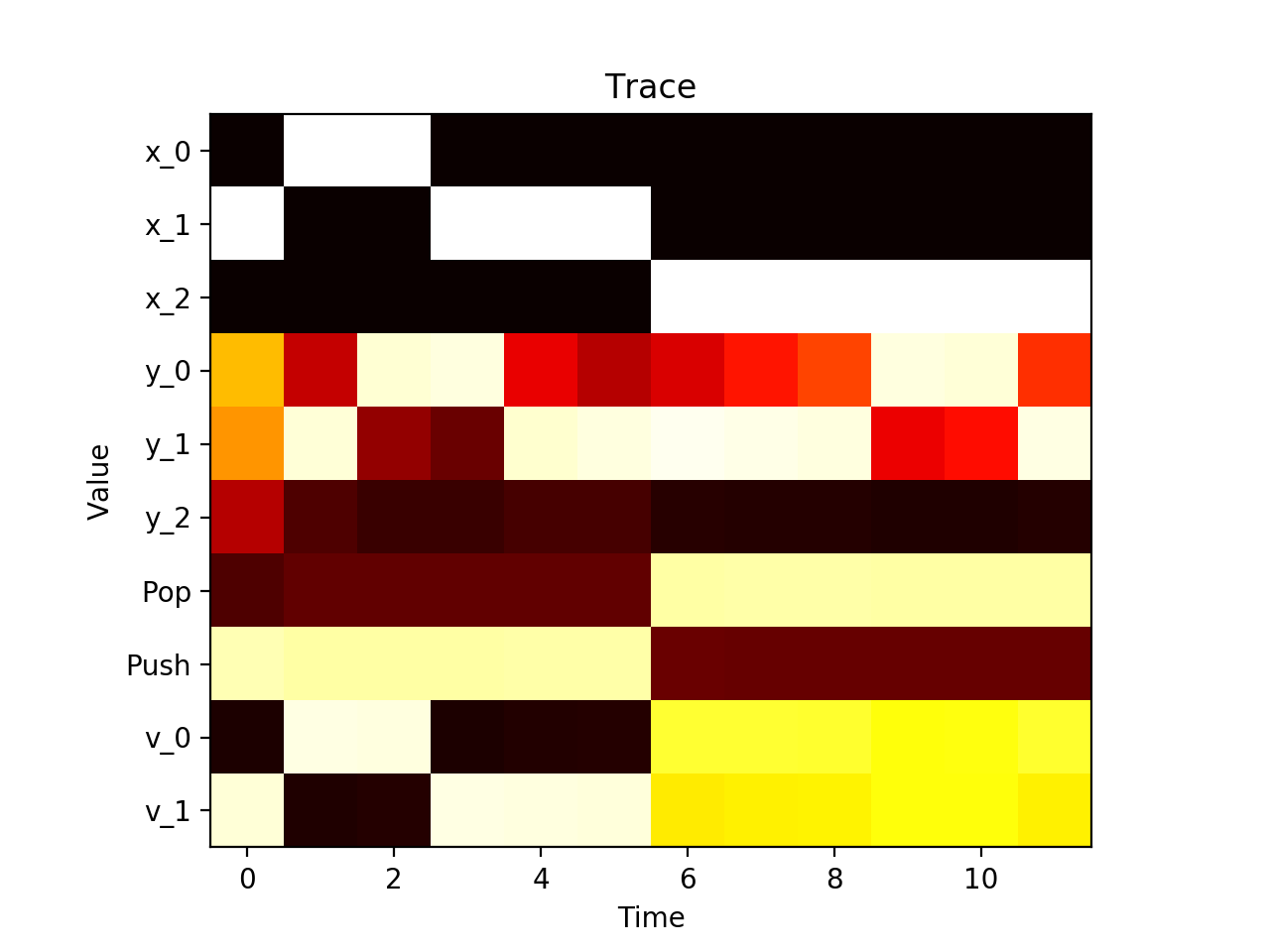} &
				\includegraphics[scale=0.25]{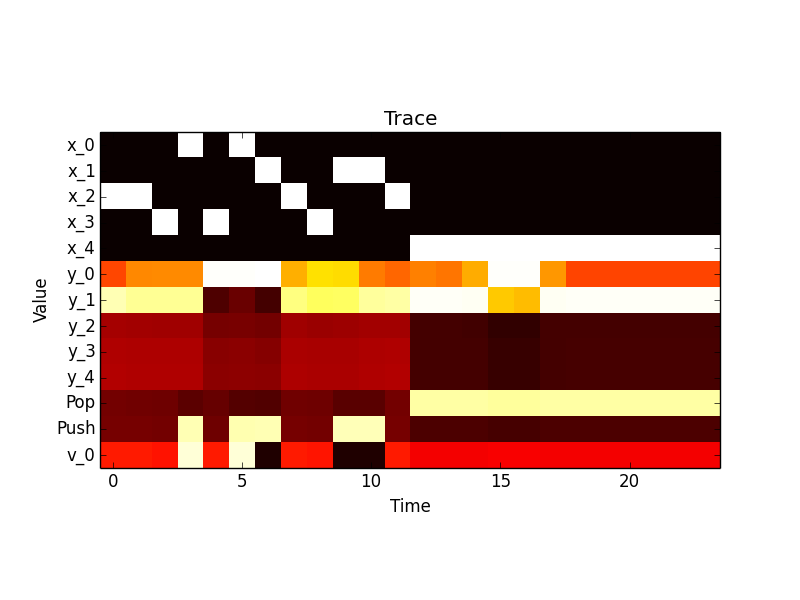} &
				\includegraphics[scale=0.3]{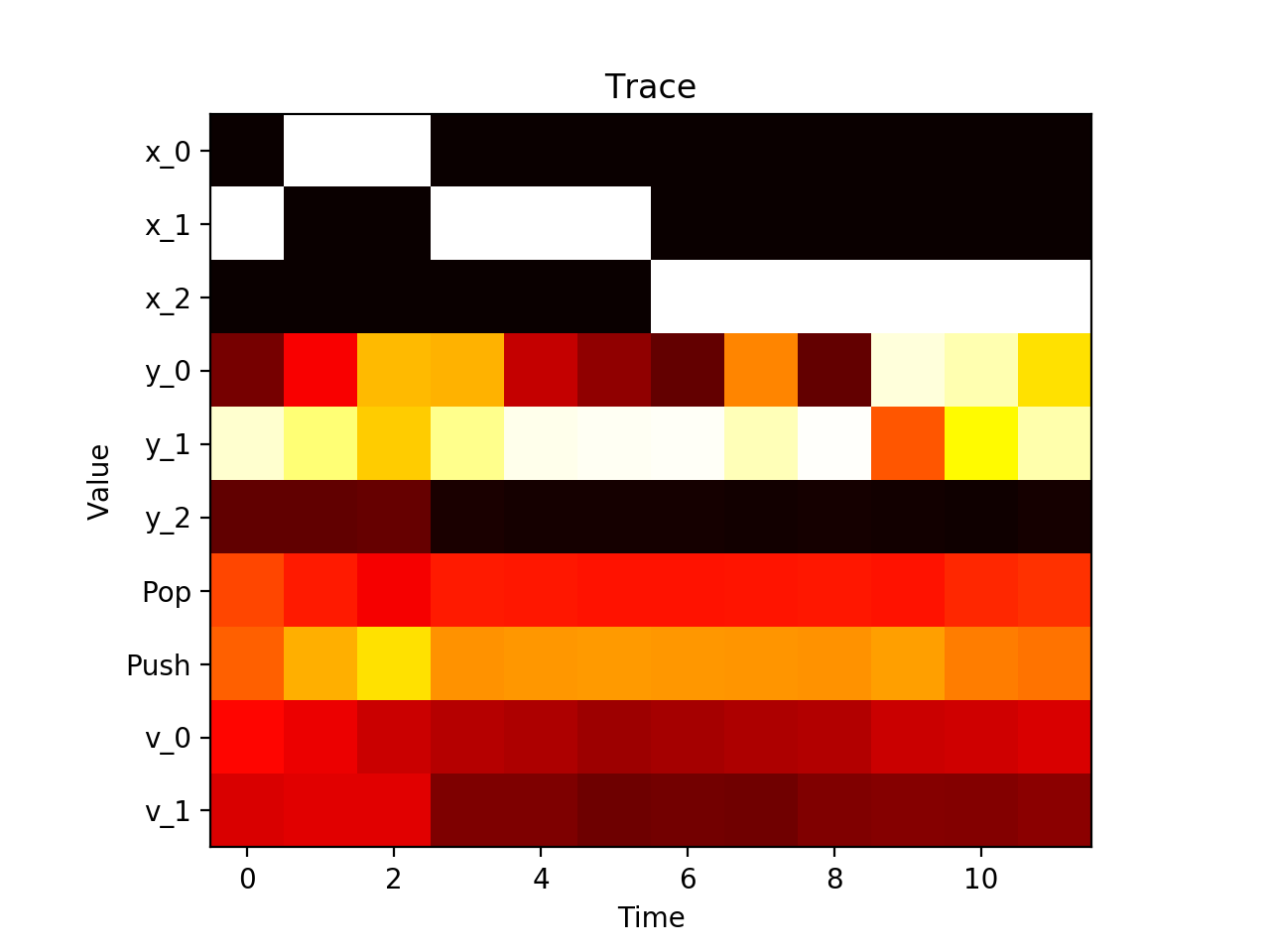} 
				 \\
				2 symbols, linear controller & 4 symbols, linear controller &  2 symbols, LSTM controller \\
				Input: $\highlight{100111}\#\ldots\#$ &
				Input: $223\highlight{0}3\highlight{01}23\highlight{11}2\#\ldots\#$ &
			    Input: $\highlight{100111}\#\ldots\#$ 
			    \\
				Output: $\ldots\highlight{111001}$ & 
				Output: $\ldots \highlight{11100}\ldots$ & 
				Output: $\ldots\highlight{111001}$ 
			\end{tabular}
			
		\end{center}
		\caption{Diagrams of network computation on the Reversal task with linear and LSTM controllers. In each diagram, the input may consist of 2 or 4 distinct alphabet symbols, but only the symbols $0$ and $1$ are included in the output. Columns indicate the pop strengths, push strengths, and pushed vectors throughout the course of the computation, along with the input and predicted output in one-hot notation. Lighter colors indicate higher values.}
	\end{figure*}
	
	\begin{figure*}[h]
		\begin{center}
			\begin{tabular}{cc}
				\includegraphics[scale=0.3]{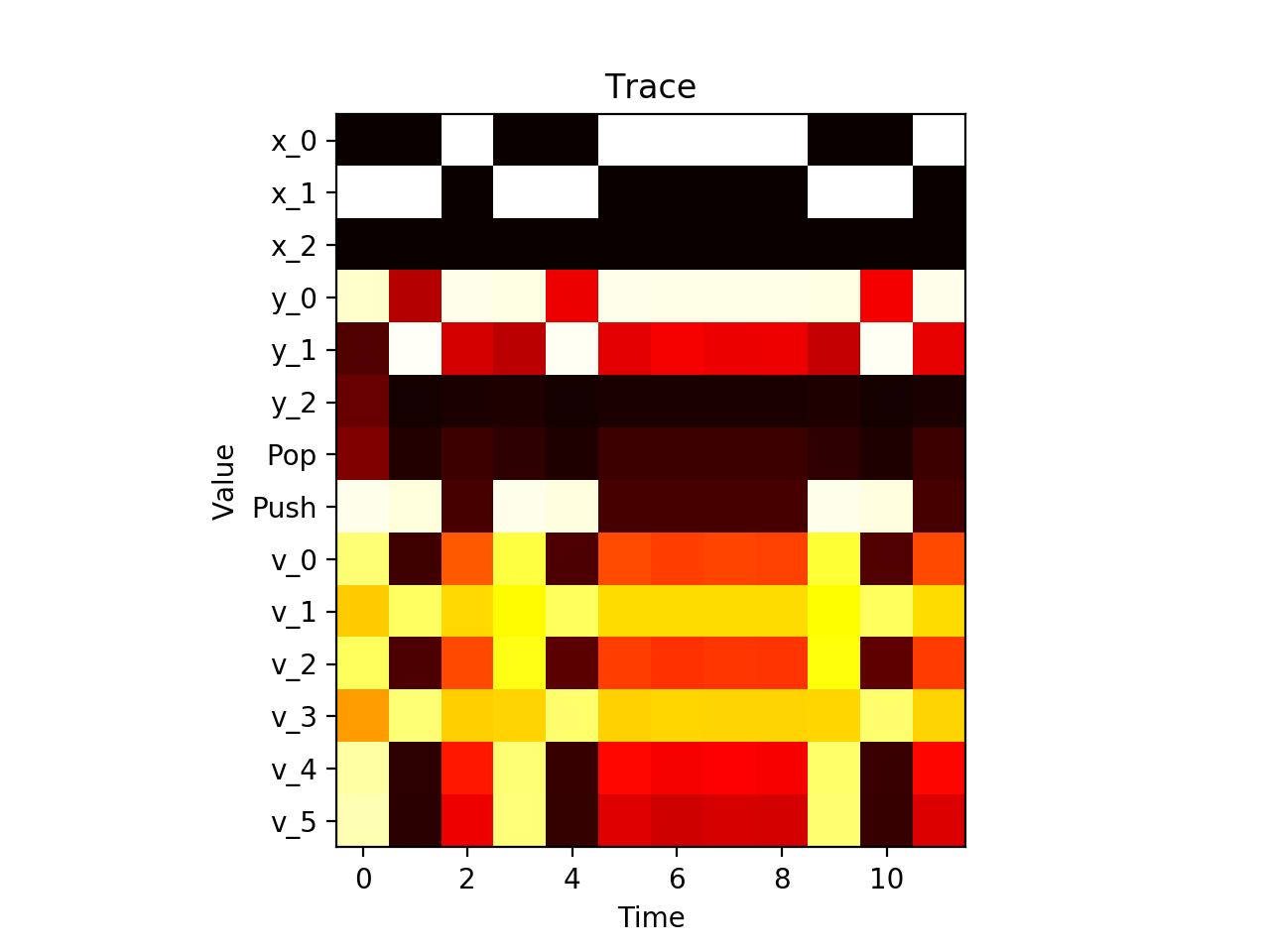} & 
				 \includegraphics[scale=0.3]{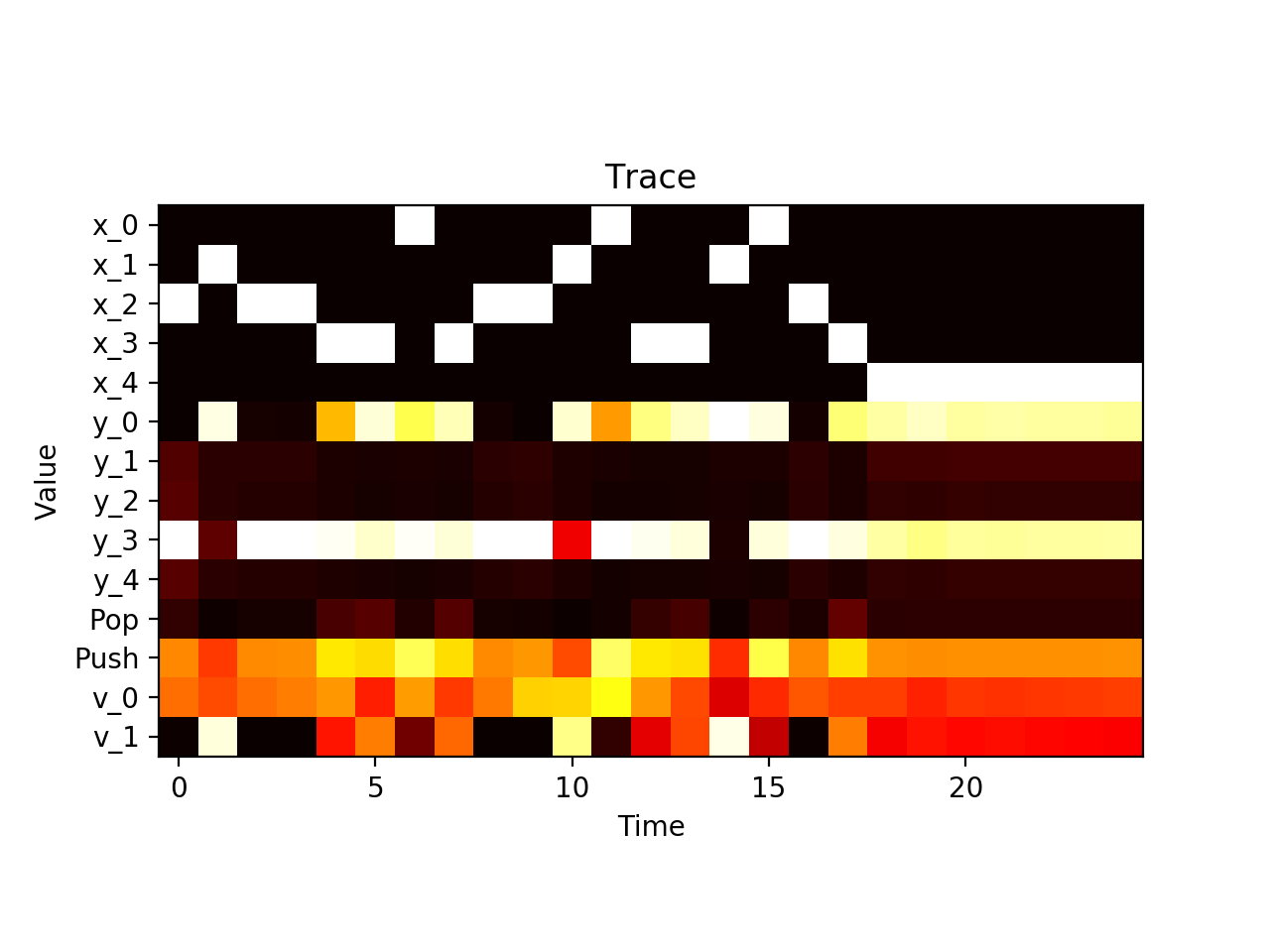} \\
				 Delayed XOR, linear controller & 
				 Parenthesis, linear controller \\
				 Input: \highlight{11011000011}0
				 & Input: [([[\highlight{]])]}[[(\highlight{)]]}(\highlight{)}[\highlight{]} 
			    \\
				Output: 0\highlight{10010000010}
				& Output: ])]\highlight{]]]]}]]]\highlight{)]]}]\highlight{)}]\highlight{]}]
			\end{tabular}
			
		\end{center}
		\caption{Diagrams of network computation for the Delayed XOR and Parenthesis tasks with a linear controller.}
	\end{figure*}
	
	\section{Results}
	
	Our results are shown in Table 1. The networks we trained were able to achieve a median accuracy of at least 90.0\% during the training phase in 10 of the 13 experimental conditions involving a stack-augmented architecture. However, many of these conditions include trials in which the model performed considerably worse during training than the median. This suggests that while stack-augmented networks are able to perform our tasks in principle, they may be more difficult to train than traditional RNN architectures. Note that there is substantially less variation in the performance of the LSTM networks without a stack.
	
	In Experiment 1, the standard network with the linear controller performs perfectly both during the training phase and in the final testing phase. The buffered network performed nearly as well during the training phase, but its performance failed to generalize to longer strings. The LSTM network achieved roughly the same performance both with and without a stack, substantially worse than the linear controller. The leftmost graphic in Figure 5 shows that the linear controller pushes a copy of its input to the stack and then pops the copy to produce the output. As suggested by an anonymous reviewer, we also considered a variant of this task in which certain alphabet symbols are excluded from the reversed output. The center graphic in Figure 5 shows that for this task, the linear controller learns a strategy in which only symbols included in the reversed output are pushed to the stack. The rightmost graphic shows that LSTM controller behaves differently from the linear controller, exhibiting uniform pushing and popping behavior throughout the computation. This suggests that under our experimental conditions, the LSTM controller prefers to rely on its recurrent state for memory rather than the stack, even though such a strategy does not scale to the final testing round.
	
	The models in Experiment 2 perform as we expected. The unbuffered model with the linear controller performed at chance, in line with the inability of the linear controller to compute XOR. The rest of the models were able to achieve accuracy above 95.0\% both in the training phase and in the final testing phase. The buffered network was successfully able to delay its computation in the Cumulative XOR Evaluation task. The leftmost graphic in Figure 6 illustrates the network's behavior in the Delayed XOR Evaluation task, and shows that the linear controller uses the stack as unstructured memory---an unsurprising observation given the nature of the task.  Note that the vectors pushed onto the stack in the presence of input symbol $1$ vary between two possible values that represent the current parity.
	
	In Experiment 3, the linear model without a stack performs fairly well during training, achieving a median accuracy of 71.8\%. This is because 43.8\% of (s and [s in the training data are immediately followed by )s and ]s, respectively, so it is possible to attain 71.9\% accuracy by predicting ) and ] when reading ( and [ and by always predicting ] when reading ) or ]. Linear models with the stack perform better, but as shown by the rightmost graphic in Figure 6, they do not make use of a stack-based strategy (since they never pop), but instead appear to use the top of the stack as unstructured memory.  The LSTM models perform slightly better, achieving 100\% accuracy during the training phase. However, the LSTM controller still suffers significantly in the final testing phase with or without a stack, suggesting that the LSTM models are not employing a stack-based strategy.
	
	In Experiment 4, the Boolean Formula Evaluation task is performed easily, with a median accuracy exceeding 90.0\% for all models both on the development set and the testing set. This is most likely because, on average, three quarters of the nodes in a boolean formula either require no context for evaluation (because they are atomic) or make use of limited context (because they are boolean formulas of depth one). The linear controller performed worse on average than the LSTM models on the agreement task, though the highest-performing linear models achieved a comparable accuracy to their LSTM counterparts. Again, the performance of the LSTM networks is unaffected by the presence of the stack, suggesting that our trained models prefer to use their recurrent state over the stack.
	
	\section{Conclusion}
	
    We have shown in Experiments 1 and 2 that it is possible in principle to train an RNN to operate a stack and input--output buffers in the intended way. There, the tasks involved have only one optimal solution: String Reversal cannot be performed without recording the string, and the linear controller cannot solve Cumulative XOR Evaluation without introducing a delay. In the other experiments, our models were able to find approximate solutions that rely on unstructured memory, and the stack-augmented LSTMs always favored such solutions over using the stack.
    
    As we saw in Experiments 3 and 4, training examples that require full usage of the stack are rare in practice, making the long-term benefits of stack-based strategies unattractive to greedy optimization. However, the usage of a stack is necessary for a general solution to all of the problems we have explored, with the exception of the XOR Evaluation tasks. While gradual improvements in performance may be obtained by optimizing the usage of unstructured memory, the discrete nature of most stack-based solutions means that finding such solutions often requires a substantial level of serendipity. Our results then raise the question of how to incentivize controllers toward stack-based strategies during training. We leave this question to future work.
	
	\bibliography{emnlp2018.bib}
	\bibliographystyle{acl_natbib_nourl}

\end{document}